\documentclass[sigconf]{acmart}
\setcopyright{rightsretained}
\acmDOI{10.475/123_4}

\acmISBN{123-4567-24-567/08/06}

\acmConference[KDD'17]{ACM SIGKDD}{August 2017}{Halifax, Nova Scotia, Canada} 
\acmYear{2017}
\copyrightyear{2017}

\acmPrice{15.00}

\usepackage{amsmath} 
\usepackage{algorithm}
\usepackage{graphicx}
\usepackage{tabularx}
\usepackage{algpseudocode}
\usepackage{multirow}
\usepackage{array}
\usepackage[caption=false]{subfig}
\usepackage{array}
\usepackage{color}
\usepackage{enumitem}
\usepackage{hhline}
\newcommand*{\affmark}[1][*]{\textsuperscript{#1}}

\newcolumntype{L}[1]{>{\raggedright\let\newline\\\arraybackslash\hspace{0pt}}m{#1}}
\newcolumntype{C}[1]{>{\centering\let\newline\\\arraybackslash\hspace{0pt}}m{#1}}
\newcolumntype{R}[1]{>{\raggedleft\let\newline\\\arraybackslash\hspace{0pt}}m{#1}}

\begin{document}
%
\title{Federated Tensor Factorization for Computational Phenotyping}

\author{Yejin Kim\affmark[1], Jimeng Sun\affmark[2], Hwanjo Yu\affmark[1], Xiaoqian Jiang\affmark[3]}\thanks{X. Jiang and H. Yu are co-corresponding authors}
\affiliation{
	\institution{\affmark[1] Pohang University of Science and Technology, Pohang, Korea}
	\institution{\affmark[2] Georgia Institute of Technology, Atlanta, GA}
	\institution{\affmark[3] University of California, San Diego, La Jolla, CA}
}
\email{(yejin89, hwanjoyu)@postech.ac.kr, jsun@cc.gatech.edu, x1jiang@ucsd.edu}


\begin{abstract}
Tensor factorization models offer an effective approach to convert massive electronic health records into meaningful clinical concepts (phenotypes) for data analysis. These models need a large amount of diverse samples to avoid population bias. An open challenge is how to derive phenotypes jointly across multiple hospitals, in which direct patient-level data sharing is not possible (e.g., due to institutional policies). In this paper, we developed a novel solution to enable federated tensor factorization for computational phenotyping without sharing patient-level data. We developed secure data harmonization and federated computation procedures based on alternating direction method of multipliers (ADMM). Using this method, the multiple hospitals iteratively update tensors and transfer secure summarized information to a central server, and the server aggregates the information to generate phenotypes.     
We demonstrated with real medical datasets that our method resembles the centralized training model (based on combined datasets) in terms of accuracy and phenotypes discovery while respecting privacy.
\end{abstract}



\renewcommand{\shortauthors}{Y. Kim et al.}

\maketitle

\section{Introduction}

Electronic health records (EHRs) become one of most important sources of information about patients, which provide insight into diagnoses~\cite{Kononenko:2001up}  and prognoses~\cite{ebadollahi_predicting_2010}, as well as assist in the development of cost-effective
treatment and management programs~\cite{Bennett:2012gc,Greengard:2013iu}. But meaningful use of EHRs is also accompanied with many challenges, for example, diversity of populations, heterogeneous of information, and data sparseness. The large degree of missing and erroneous records also complicates the interpretation and analysis of EHRs. 
Furthermore, clinical scientists are not used to the complex and high-dimensional EHR data ~\cite{Cimino:2000uh,Ledley:1991wc}. Instead, they are more accustomed to reasoning based on accurate and concise clinical concepts (or phenotypes) such as diseases and disease subtypes. Useful phenotypes should capture multiple
aspects of the patients (e.g., diagnosis, medication and lab results)
and be both sensitive and specific to the target patient population. 
Although some phenotypes can be easily concluded based on EHR data, a wide range of clinically important ones such as disease subtypes  are not obtainable in a straightforward manner. 
The transformation from EHR data into useful phenotypes, or \emph{phenotyping} is a fundamental challenge to learn from EHR data. Current approaches for translating EHR data into useful phenotypes are typically slow, manually intensive
and limited in scope~\cite{Carroll:2011ue,Carroll:2012jr}. 
Overcoming several disadvantages of the previous methods, tensor factorization methods have shown great potential in discovering meaningful phenotypes from complicated and heterogeneous health records~\cite{ho2014limestone, ho_marble:_2014, wang2015rubik}. 

Nevertheless, phenotypes developed from one hospital are often limited due to a small sample size and inherent population bias. Ideally, we would like to compute phenotypes on a large population with data combined from multiple hospitals. However, this will require healthcare data sharing and exchange, which are often impeded by policies due to the privacy concerns. For example, PCORnet data privacy guidance does not allow record-level research participant information sharing and it recommends a minimum count threshold (e.g., 10) for aggregate data sharing \cite{PCORnetDataPrivacyTaskForce2015}. The same threshold is used in Informatics for Integrating Biology \& the Bedsides (I2B2) \cite{murphy2010serving}, a famous system developed by National Center for Biomedical Computing based at Partners HealthCare. The real-world challenges motivate the development of a {\it federated phenotyping method} to learn phenotypes across multiple hospitals  with mitigated privacy risks.  

In the federated method, the hospitals perform most of computations, and a semi-trusted server supports the hospital by aggregating results from hospitals.
The hospitals demand a certain form of summarized patient information (not patient-level data) anyhow for updating tensor. 
A challenge of the federated tensor factorization is that the summarized information can disclose the patient-level data.
For example, an objective function of tensor factorization is $||\mathcal{X-O}||^2$ where $\mathcal{X}$ is a tensor to be estimated using an observed tensor $\mathcal{O}$.
Because the objective function is not linearly separable over hospitals, tensor factorization for each hospital inevitably demands the others patient-level data.
Thus, hospitals should share summarized information that does not disclose the patient-level data but instead contains accurate phenotypes from the patient-level data.

However, sharing the summarized information raises another challenge when the data are distributed in many hospitals as a relatively small size, or when the data are unevenly distributed. 
Because of sampling error, noise in the summarized information can increase with small patient populations. Accuracy then can be decreased or unstable. 
Therefore, we need to ensure the robustness of summary information even with small sized or unevenly distributed samples. 

In this paper, we develop federated \textbf{T}ensor factorization for p\textbf{ri}vacy preserving computational \textbf{p}henotyping (\textsc{Trip}), a new federated framework for tensor factorization over horizontally partitioned data (i.e., data are partitioned based on rows or patients).
Our major contributions are the following:

\textbf{i) Accurate and fast federated method}: \textsc{Trip} is as accurate as centralized training model (based on combined datasets). The accuracy of \textsc{Trip} is robust on the patient size or distribution. \textsc{Trip} is fast compared to the centralized training model thanks to federated computation.  

\textbf{ii) Rigorous privacy and security analysis}: \textsc{Trip} preserves the privacy of patient data by transferring summarized information. We prove that the summarized information does not disclose the patient data. 

\textbf{iii) Phenotype discovery from real datasets}: Phenotypes that \textsc{Trip} discovers without sharing the patient-level data are the same phenotypes based on the combined data. \textsc{Trip} even discovers some phenotypes that individual hospital cannot discover due to biased and limited population. 

\section{RELATED WORKS}
Many privacy preserving data mining algorithms aim at constructing a global model only from aggregated statistics locally generated by participating institutions on their own data, without seeing others' data at a fine-grained level \cite{vaidya2008privacy, li2015vertical}. More rigorous privacy criteria like \textit{differential privacy} \cite{dwork06}, which introduces noises, have been applied for several classification models through parameter or objective perturbations \cite{CMS11}. However, this is not desirable for computational phenotyping applications because noise can lead to ``ghost'' phenotypes, which do not exist in the original databases and might mislead healthcare providers with severe consequences. In this work, we will consider privacy protection like in the former privacy preserving data mining methods to compute phenotypes by only exchanging summary statistics, calculated by local participants.

Tensor factorization emerged as a promising solution for computational phenotyping thanks to its interpretability and flexibility. In the medical context, tensor factorization has been adapted to enforce sparsity constraints \cite{ho2014limestone}, model interactions among groups of the same modality \cite{ho_marble:_2014}, and absorbing prior medical knowledge via customized regularization terms \cite{wang2015rubik}. Our goal is to develop a federated tensor factorization framework to compute phenotypes in a privacy-preserving way. This is different from distributed tensor factorization models \cite{choi2014dfacto, kang2012gigatensor} and grid tensor factorization models \cite{de2014distributed}. The latter assumes data spread across different but interconnected computer systems, in which the communication cost is negligible and data/computation can be arbitrarily reallocated to improve parallelization efficiency. In contrast, our \textsc{Trip} framework deals with data stored in separate sources (hospital at different locations) and requires the ability to go through policy barriers using accepted practices that respect privacy.

\begin{table}[t]
\centering
\caption{Notations and symbols}
\vspace{-3mm}
\label{notation}
\begin{tabular}{|l|l|}
\hline
$\circ$                & outer product                         \\ \hline
$\odot$                & Khatri-Rao product                         \\ \hline
$R$                    & number of ranks  \\ \hline
$N$                    & number of modes (order)                        \\ \hline
$K$                    & number of hospitals                 \\ \hline
$\mathbf{A, B}$        & matrix                                \\ \hline
$\mathcal{X, O}$ & tensor                                \\ \hline
$\mathbf{O}_{(n)}$        & matricized tensor of $\mathcal{O}$ on $n$th mode                   \\ \hline
\end{tabular}
\vspace{-3mm}
\end{table}

\section{PRELIMINARIES}\label{sec:tensorFactorization}
We first describe some preliminaries of tensor factorization, and summarize the notations and symbols in Table \ref{notation}. 

\begin{definition}
Outer product of $N$ vectors $\mathbf{a}^{(1)} \circ \cdots \circ \mathbf{a}^{(N)} $ makes $N$-order rank-one tensor $\mathcal{X}$. 
\end{definition}
\begin{definition}
Kronecker product of two vectors $\mathbf{a} \in \mathbb{R}^{I_a \times 1}$ and $\mathbf{b} \in \mathbb{R}^{I_b \times 1}$ is
\begin{equation*}
\mathbf{a}  \otimes \mathbf{b} = 
\begin{bmatrix}
a_1 \mathbf{b}\\
\vdots \\
a_{I_a} \mathbf{b}
\end{bmatrix}
\in \mathbb{R}^{I_a I_b \times 1}.
\end{equation*}
\end{definition}
\begin{definition}
Kharti-Rao product of two matrices $\mathbf{A} \in \mathbb{R}^{I_A \times R}$ and $\mathbf{B} \in \mathbb{R}^{I_B \times R}$ is  $\mathbf{A} \odot \mathbf{B}= \left[ \mathbf{a}_1 \otimes \mathbf{b}_1\cdots  \mathbf{a}_R \otimes \mathbf{b}_R \right] \in \mathbb{R}^{I_A I_B \times R}$.
\end{definition}

\begin{definition}
\textit{Matricization} is to reshape the tensor into a matrix by unfolding elements of the tensor. Mode-$n$ matricization of tensor $\mathcal{O}$ is denoted as $\mathbf{O}_{(n)}$. 
\end{definition}

Tensor factorization is a dimensionality reduction approach that represents the original tensor as a lower dimensional latent matrix. The CANDECOMP/PARAFAC (CP) \cite{carroll1970analysis} model is the most common tensor factorization, which approximates the original tensor $\mathcal{O}$ as $\mathcal{X}$, a linear combination of $R$ rank-one tensors that are made from outer product of $N$ vectors. That is, CP tensor factorization is represented as
\begin{equation*}
\begin{split}
\mathcal{O} \approx \mathcal{X} 
= \sum_{r=1}^R \mathbf{A}^{(1)}(: ,r) \circ \cdots \circ \mathbf{A}^{(N)}(: ,r),
\end{split}	
\end{equation*}
where 
$\mathbf{A}^{(n)}(: ,r)$ refers to the $r$th column of $\mathbf{A}^{(n)}$. Here, $\mathbf{A}^{(n)}$ is the $n$th factor matrix. 
$R$ is referred as the rank of the $\mathcal{X}$.   
The columns from factor matrices represent latent concepts that describe the data as lower dimensions. 


Tensor factorization for phenotyping is to compute a factorized tensor $\mathcal{X}$ that contains latent medical concepts from data (or observed tensor $\mathcal{O}$).
 $\mathcal{X}$ consists of the $R$ most prevalent phenotypes. The $n$th factor matrix, $\mathbf{A}^{(n)}$ defines the elements from the mode $n$ to comprise the phenotypes. That is, $r$th phenotype consists of $r$th column of factor matrices \cite{ho2014limestone}. 

The objective function of the tensor factorization with regularization terms for pairwise distinct constraints \cite{wang2015rubik} is formulated as 
\begin{equation}
\begin{split}
\min_ {\mathcal{X}} \Psi =&||\mathcal{X-O}||^2_F +\sum_{n=1}^N \frac{\lambda}{2}||\mathbf{I-A}^{(n)T}\mathbf{A}^{(n)} ||^2_F.
\end{split}
\label{eq:obj}
\end{equation}
It is rewritten with respect to mode-$n$ matricization
\begin{equation}
\min_ {\mathbf{A}^{(n)}} \Psi = || \mathbf{A}^{(n)}\Pi^{(n)T} -\mathbf{O}_{(n)} ||^2_F + \frac{\lambda}{2}||\mathbf{I-A}^{(n)T}\mathbf{A}^{(n)} ||^2_F.
\end{equation}
where
$\Pi^{(n)}= \mathbf{A}^{(N)}\odot  \ldots \odot \mathbf{A}^{(n+1)} \odot \mathbf{A}^{(n-1)} \odot \ldots \odot \mathbf{A}^{(1)}$.
This is our decomposition goal in the rest of this paper. Solving the problem (\ref{eq:obj}) while preserving privacy is technically challenging because the tensor residual term $\mathcal{X-O}$ inherently contains other hospitals' patient data that involve sensitive information. 

\section{Federated Tensor Factorization}
We first provide a general overview of the \textsc{Trip} and then formulate the problem with iterative updating rules for optimization.
\subsection{Overview}
\begin{figure}[t]
\centering
\includegraphics[width=0.45\textwidth]{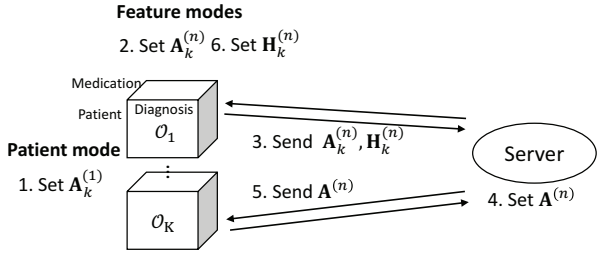}
\caption{Process of federated tensor factorization.}
\label{fig:process}
\end{figure}

\begin{figure}[t]
\centering
\includegraphics[width=0.45\textwidth]{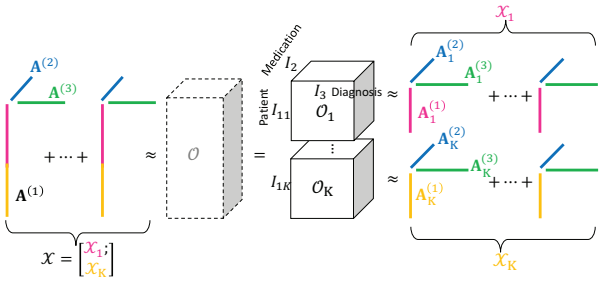}
\caption{Equivalence between tensor factorization with respect to each local tensor $\mathcal{O}_k$ and tensor factorization with respect to global tensor $\mathcal{O}$. Without $\mathcal{O}$, tensor factorization that is globally optimal across hospitals can be achieved via local tensor factorization.}
\label{fig:decomposition}
\end{figure}

\textsc{Trip} is a federated tensor factorization for horizontally partitioned patient data. We assume the data are horizontally partitioned along patient mode, that is, hospitals have their own patient data on the same medical features (Figs. \ref{fig:process}, \ref{fig:decomposition}).
Let us assume that there are $K$ hospitals and a central server, where the server distributes most decomposition computation to hospitals and aggregates intermediate results from them. We assume \emph{Honest-but-Curious} adversary model, in which the server and hospitals are curious on data of others but do not maliciously manipulate intermediate results \cite{kissner2005privacy}. 

A local observed tensor $\mathcal{O}_k$ is the local patient data in hospital $k$ (Fig. \ref{fig:decomposition}); a local factorized tensor $\mathcal{X}_k$ is the factorized tensor generated by local observed tensor in hospital $k$,
$\mathcal{X}_k$ has $N$ modes for the set of patient and medical features (eg. medication, diagnosis). In this case, $N=3$ because we have modes for patient, medication, and diagnosis. 
The horizontally partitioned patient mode of each $\mathcal{X}_k$ is generated from distinct set of patients whose size is $I_{1k}$.
For simplicity, first mode ($n=1$) always denotes patient mode.
On the other hand, $N-1$ medical features modes that hospitals share of each $\mathcal{X}_k$ is generated from the same set of $N-1$ medical features whose size is $I_n$,  ($n= 2, \ldots, N$).
For example, diagnosis and medication can be the feature modes. The size of $\mathcal{X}_k$ is ${I_{1k} \times I_2 \times \cdots \times I_N}, ~\forall k$.

We assume that factor matrix on feature modes of the local factorized tensor $\mathcal{X}_k$ is the same for all the hospitals. By assuming that, all hospitals are enforced to share the same phenotypes. Also, the objective function $\Psi$ in Eq. (\ref{eq:obj}) can be linearly separable on hospitals; consequently hospitals can update their local factorized tensor indirectly using other hospitals' patient data while respecting privacy. 

The local factorized tensor $\mathcal{X}_k$ is computed as following steps: first, in patient mode, hospital $k$ ($k=1, \ldots, K$) computes local factor matrix independently (step 1) in Fig. \ref{fig:process}. 
For feature modes, hospital $k$ computes the local factor matrices (step 2) and send them together with the Lagrangian multipliers to the server (step 3). The server then generates harmonized factor matrix (global factor matrix) by combining all the local factor matrices with Lagrangian multipliers (step 4). After receiving the global factor matrix (step 5), hospital $k$ updates the Lagrangian multipliers (step 6). 
Hospitals and the server repeat the procedures until the local factor matrices are converged. 
During the procedures, the global factor matrices can retain phenotypes from local factor matrices without directly using the local patient data.

\subsection{Formulation}\label{sec:formulation}
We first formulate separable objective function on hospitals for federated tensor factorization. 
The objective function for tensor factorization, $\Psi$ in Eq. (\ref{eq:obj}) is reformulated with respect to the local factorized tensor.

$\mathcal{X}_k$ is decomposed into factor matrices  $\mathbf{A}_k^{(1)} \in \mathbb{R}^{I_{1k} \times R}$ (patient mode) and $\mathbf{A}_k^{(n)} \in \mathbb{R}^{I_n \times R},  n \geq 2 $ (feature modes).
We assume that the local factor matrices of feature modes $\mathbf{A}_k^{(n)}$ from all hospitals are equal to the global factor matrix (Fig. \ref{fig:decomposition}), i.e.,
\begin{equation}
\mathbf{A}^{(n)}=\mathbf{A}_1^{(n)} =\mathbf{A}_2^{(n)} = \ldots = \mathbf{A}_K^{(n)}, ~~ n \geq 2.
\label{eq:assumption}
\end{equation}
This assumption is reasonable because all hospitals aim to have the same phenotypes and share them with others. 
By assuming Eq. (\ref{eq:assumption}), the horizontal concatenation of the local factor matrices of patient mode $\mathbf{A}_k^{(1)}$ forms the (global) factor matrix $\mathbf{A}^{(1)}$ (Fig. \ref{fig:decomposition}):
\begin{equation}
\mathbf{A}^{(1)}=
\begin{bmatrix}
\mathbf{A}_1^{(1)}; \\
 \vdots \\
 \mathbf{A}_K^{(1)}
\end{bmatrix}.
\end{equation}
Accordingly, we represent the global factorized tensor $\mathcal{X}$ in Eq. (\ref{eq:obj}) with respect to the local factorized tensor $\mathcal{X}_k$ (Fig. \ref{fig:decomposition}) as 
\begin{equation*}
\begin{split}
\mathcal{X} =
\begin{bmatrix}
\mathcal{X}_1 ;\\
\vdots\\
\mathcal{X}_K \\
\end{bmatrix}
=
\begin{bmatrix}
 \sum_r \mathbf{A}_1^{(1)}(:, r) \circ \mathbf{A}^{(2)}(:,r) \circ \cdots \circ \mathbf{A}^{(N)}(:, r) ;\\
\vdots\\
\sum_r \mathbf{A}_K^{(1)}(:,r) \circ \mathbf{A}^{(2)}(:,r) \circ \cdots \circ \mathbf{A}^{(N)}(:, r) \\
\end{bmatrix},
\label{eq:T}
\end{split}
\end{equation*}
and we can make the objective function $\Psi$ linearly separable on $k$ as
$||\mathcal{X-O}||^2_F = \sum_{k=1}^{K}||\mathcal{X}_k-\mathcal{O}_k||^2_F$.
The optimization problem for tensor factorization is reformulated with respect to local tensors:
\begin{equation}\label{eq:optimizationProblem}
\begin{split}
\min_{\mathcal{X}_k}  \Psi = \sum_{k=1}^{K}&||\mathcal{X}_k-\mathcal{O}_k||^2_F + \sum_{n=2}^{N}
\frac{\lambda}{2}||\mathbf{I-B}^{(n)T}\mathbf{A}^{(n)} ||^2_F \\
s.t. 
&\mathbf{A}^{(n)}= \mathbf{A}_k^{(n)} ~~ n \geq 2, \forall k \\
&\mathbf{B}^{(n)}=\mathbf{A}^{(n)} ~~  n \geq 2.
\end{split}
\end{equation}
Here, the non-convex second term $||\mathbf{I-A}^{(n)T}\mathbf{A}^{(n)} ||^2_F$ in Eq. (\ref{eq:obj}) is replaced to a convex term  $||\mathbf{I-B}^{(n)T}\mathbf{A}^{(n)} ||^2_F$ using $\mathbf{B}^{(n)}$ such that $\mathbf{A}^{(n)} =\mathbf{B}^{(n)}$.
We assume that the pairwise constraint is only applied to the feature modes.
This assumption is reasonable because phenotypes are defined as only combination of medical features in feature modes.

Augmented Lagrangian function $\mathcal{L}$ for the new optimization problem (\ref{eq:optimizationProblem}) is
\begin{equation*}
\begin{split}
\mathcal{L}=\Psi +&\sum_{k=1}^K \sum_{n=2}^N \left[( \mathbf{A}^{(n)}-\mathbf{A}_k^{(n)})^T \mathbf{H}_k^{(n)}  + \frac{\omega }{2}||  \mathbf{A}^{(n)}- \mathbf{A}_k^{(n)}||^2_F \right] \\
+& \sum_{n=2}^N \left[ (\mathbf{B}^{(n)}-\mathbf{A}^{(n)})^T\mathbf{Y}^{(n)}  + \frac{\mu}{2}||  \mathbf{B}^{(n)}-\mathbf{A}^{(n)} ||^2_F  \right]\\
\end{split}
\end{equation*}
where $\mathbf{H}_k^{(n)}$ and $\mathbf{Y}^{(n)}$ are the Lagrangian multipliers. The penalty terms that are multiplied by parameter $\omega$ and $\mu$ help $\mathcal{L}$ to improve the convergence property (i.e.,\textit{ method of multiplier}) \cite{lin2011linearized} during federated optimization in Section \ref{sec:distributedOptimization}.

\subsection{Federated optimization}\label{sec:distributedOptimization}
The optimization problem (\ref{eq:optimizationProblem}) is then solved via consensus alternating direction method of multipliers (ADMM) \cite{boyd2011distributed}, which decomposes the original problem into sub-problems using auxiliary variables and ensures convergence to a stationary point even with nonconvex problem \cite{hong2016convergence}. 
Our problem is decomposed to sub-problems for hospitals with respect to the local factor matrices. Individual components of the local factor matrices are iteratively updated while other local factor matrices are fixed.
Once all hospitals update the local factor matrices, server updates the global factor matrix and send it back to every hospital. 
Hospitals and server repeat this procedure until the local factor matrices converge before maximum iteration.

\subsubsection{Patient mode}
Because the factor matrix for patient mode does not need to be shared, each hospital updates the local factor matrix without sharing the intermediate results. 
The local matricized residual tensor on patient mode is 
\begin{equation*}
\begin{split}
(\mathcal{X}_k - \mathcal{O}_k)_{(1)}& = \mathbf{A}_k^{(1)}\Pi^{(1)T} - \mathbf{O}_{(1)k} \in \mathbb{R}^{I_{1k} \times (I_2 \cdots I_N)}.
\end{split}
\end{equation*}
Horizontal concatenation of the local matricized residual tensors  $\mathbf{A}_k^{(1)}\Pi^{(1)T} - \mathbf{O}_{(1)k}$ from $K$ hospitals becomes the global matricized residual tensor $\mathbf{A}^{(1)}\Pi^{(1)T} -\mathbf{O}_{(1)}$.
To compute $\mathbf{A}_k^{(1)}$, we separate the first term in $\Psi$ in Eq. (\ref{eq:obj}) to each hospital as
\begin{equation}
||\mathbf{A}^{(1)}\Pi^{(1)T} -\mathbf{O}_{(1)}||_F^2 =
\sum_{k=1}^K ||\mathbf{A}_k^{(1)}\Pi^{(1)T} -\mathbf{O}_{(1)k}||_F^2.
\end{equation}
By setting derivatives of $\Psi$ with respect to $\mathbf{A}_k^{(1)}$ to zero,
a closed form solution for updating $\mathbf{A}_{k}^{(1)}$ is
\begin{equation}
\mathbf{A}_{k}^{(1)}=\{  \mathbf{O}_{(1)k}\Pi^{(1)}\}\{ \Pi^{(1)T}\Pi^{(1)} \}^{-1} .
\label{eq:Ai1}
\end{equation}

\subsubsection{Feature modes}
Hospitals update the local factor matrices using the global factor matrix, and server makes the global factor matrix by aggregating the intermediate local factor matrices from hospitals in turn. 
\\
\textbf{Update the local factor matrices:} 
The local matricized residual tensor on feature modes is 
\begin{equation*}
(\mathcal{X}_k - \mathcal{O}_k)_{(n)} = \mathbf{A}^{(n)}\Pi_k^{(n)T} - \mathbf{O}_{(n)k} \in \mathbb{R}^{I_n \times ( I_N \cdots I_{n+1} I_{n-1} \cdots I_{1k})}
\end{equation*}
where 
$\Pi_k^{(n)}= \mathbf{A}_{k}^{(N)}\odot  \ldots \odot \mathbf{A}_{k}^{(n+1)} \odot \mathbf{A}_{k}^{(n-1)} \odot \ldots \odot \mathbf{A}_{k}^{(1)}, ~ n  \geq 2$.
Contrast to patient mode, vertical concatenation of the local matricized residual tensors $\mathbf{A}^{(n)}\Pi_k^{(n)T} - \mathbf{O}_{(n)k}$ becomes the global matricized residual tensor $\mathbf{A}^{(n)}\Pi^{(n)T} -\mathbf{O}_{(n)}$.
The first term in $\Psi$ becomes
\begin{equation}
||\mathbf{A}^{(n)}\Pi^{(n)T} -\mathbf{O}_{(n)}||_F^2= \sum_{k=1}^K ||\mathbf{A}_k^{(n)}\Pi_k^{(n)T} - \mathbf{O}_{(n)k}||_F^2
\end{equation}
with $\mathbf{A}^{(n)} =\mathbf{A}_k^{(n)}$.
The closed form solution for $\mathbf{A}_{k}^{(n)}$ is
\begin{equation}
\mathbf{A}_{k}^{(n)}=\{  \mathbf{O}_{(n)k}\Pi_k^{(n)}+\omega \mathbf{A}^{(n)}+\mathbf{H}_{k}^{(n)} \}\{ \Pi_k^{(n)T}\Pi_k^{(n)} +\omega \mathbf{I}  \}^{-1} .
\label{eq:Ain}
\end{equation}
This closed form solution updates the local factor matrices using the both local observed tensor $\mathbf{O}_{(n)k}$ and global factor matrix $\mathbf{A}^{(n)}$.
That is, each hospital uses both their patient data and the common phenotypes from others to update their local phenotypes.
Now, hospitals send the local information $\mathbf{A}_{k}^{(n)}$ and $\mathbf{H}_{k}^{(n)}$ to server for following updates on the global factor matrix. \\
\textbf{Update the global factor matrix:} 
Server updates the global factor matrix based on the local information.
The objective function is
\begin{equation*}
\begin{split}
\min_{\mathbf{A}^{(n)}} &\frac{\lambda}{2}|| \mathbf{I}-\mathbf{B}^{(n)T}\mathbf{A}^{(n)}||^2_F 
+\frac{\mu}{2} ||\mathbf{A}^{(n)}-\mathbf{B}^{(n)}-\mathbf{Y}^{(n)}/\mu ||^2_F \\
 + &\frac{\omega}{2}\sum_{k=1}^K||\mathbf{A}^{(n)}-\mathbf{A}_{k}^{(n)}+\mathbf{H}_{k}^{(n)}/\omega||^2_F
\end{split}
\end{equation*}
that also uses the pairwise constraint.
$\mathbf{A}^{(n)}$ is updated to be similar with $\mathbf{A}_{k}^{(n)}$ in the third term. That is, the global phenotypes are made to be similar with all other hospitals' phenotypes. 
By derivatives of this function with respect to $\mathbf{A}^{(n)}$, we derive the following closed form solution:
\begin{equation}
\begin{split}
\mathbf{A}^{(n)}=&\{ (\mu+K\omega)\mathbf{I} + \lambda \mathbf{B}^{(n)}\mathbf{B}^{(n)T} \}^{-1}  \\
\cdot &\{ \lambda \mathbf{B}^{(n)} +\mu \mathbf{B}^{(n)} + \mathbf{Y}^{(n)} + \omega \sum_k \mathbf{A}_{k} ^{(n)} - \sum_k \mathbf{H}_{k}^{(n)} \}.
\end{split}
\label{eq:Aw}
\end{equation}
%
Now, server sends the global information $\mathbf{A}^{(n)}$ to hospitals for the next iteration. 
Server updates $\mathbf{B}^{(n)}$ by 
\begin{equation}
\mathbf{B}^{(n)} = \mathbf{A}^{(n)} + \frac{1}{\mu} \mathbf{Y}^{(n)}.
\label{eq:B}
\end{equation}
\\
\textbf{Update Lagrangian multipliers:}
Finally, server updates Lagrangian multipliers as
\begin{equation}
\mathbf{Y}^{(n)}=\mathbf{Y}^{(n)}+\mu (\mathbf{B}^{(n)} - \mathbf{A}^{(n)}). \\
\label{eq:Y}
\end{equation}
Hospitals also updates local Lagrangian multipliers as 
\begin{equation}
\mathbf{H}_{k}^{(n)}=\mathbf{H}_{k}^{(n)}+\omega ( \mathbf{A}^{(n)} -  \mathbf{A}_{k}^{(n)} ) \\
\label{eq:H}
\end{equation}
to adjust the gap between local and global factor matrices.
The entire procedures of updating the tensors are summarized in Algorithm \ref{alg}.
\begin{algorithm}[t]
\caption{\textsc{Trip}}
\begin{algorithmic}[1]
\State \textbf{Input}: $\mathcal{O}, \lambda, \omega, \mu $ 
\State     Initialize $\mathbf{A}_{k}^{(n)},\mathbf{H}_{k}^{(n)}, \mathbf{Y}^{(n)}$.
\Repeat
	\State // Update patient mode $n=1$
	\State Hospitals set $\mathbf{A}_{k}^{(1)}~~\forall k$  \textcolor{red}{(Eq. \ref{eq:Ai1})}. 
	\For{$n=2, \ldots, N$} //Update feature modes 
		\State Hospitals set and send $\mathbf{A}_{k}^{(n)} ~~\forall k$ \textcolor{red}{(Eq. \ref{eq:Ain})}.
		\State Server sets and sends $\mathbf{A}^{(n)}$  \textcolor{red}{(Eq. \ref{eq:Aw})}.
		\State Server sets $\mathbf{B}^{(n)}$ and $\mathbf{Y}^{(n)}$ \textcolor{red}{(Eq.\ref{eq:B}, \ref{eq:Y})}.
		\State Hospitals set and send $\mathbf{H}_{k}^{(n)} ~~\forall k$  \textcolor{red}{(Eq. \ref{eq:H})}.	

		 \EndFor
\Until{Converged}
\end{algorithmic}
\label{alg}
\end{algorithm}

\subsection{Convergence proof}
We prove that our federated tensor factorization (\ref{eq:optimizationProblem}) converges.
Due to limited space, detailed proof of inequality (\ref{eq:vk}) and (\ref{eq:lowerbound}) can be found in our technical report \cite{ykim16trip} or \cite{boyd2011distributed}.
For each $n=2, \cdots, N$, let us denote 
\begin{equation}
\begin{split}
x &=[\mathbf{A}^{(n)}_1 (:), \ldots, \mathbf{A}^{(n)}_K (:)] \in \mathbb{R}^{I_n R \times K},\\
z &= [\mathbf{A}^{(n)} (:), \ldots, \mathbf{A}^{(n)} (:)] \in \mathbb{R}^{I_n R \times K},\\
y &= [\mathbf{H}^{(n)}_1 (:), \ldots, \mathbf{H}^{(n)}_K (:)] \in \mathbb{R}^{I_n R \times K}, \text{~and}\\ 
r^t & = x^t - z^t,
\end{split}
\end{equation}
for vectorized local factor matrices, global factor matrix, Lagrangian multipliers, and residual at iteration $t$, respectively.
Then $\mathcal{L}$ is rewritten as
\begin{equation}
\mathcal{L} (x, z, y) = f(x) + g(z) + y^T(x-z) + (\omega/2) ||x-z||^2 
\end{equation}
where $f(x) =\sum_{k=1}^K ||\mathbf{A}_k^{(n)} \Pi_k^{(n)T} - \mathbf{O}_{(n)k}||^2$ and $g(z) = \frac{\lambda}{2} ||\mathbf{I}- \mathbf{B}^{(n)T} \mathbf{A}^{(n)}||^2 +  (\mathbf{B}^{(n)}-\mathbf{A}^{(n)})^T\mathbf{Y}^{(n)}  + \frac{\mu}{2}||  \mathbf{B}^{(n)}-\mathbf{A}^{(n)} ||^2 $. Let $(x^*, z^*, y^*)$ be a saddle point, and define
\begin{equation}
V^t = (1/ \omega) || y^t -y^*||^2 + \omega ||z^t - z^*||^2.
\end{equation}
$V^t$ decreases in each iteration (proof in \cite{ykim16trip}) because
\begin{equation}
V^{t+1} \leq V^t - \omega || r^{t+1} ||^2 - \omega || z^{t+1} - z^t ||^2.
\label{eq:vk}
\end{equation}
Adding the inequality (\ref{eq:vk}) through $t=0$ to $\infty$ gives
\begin{equation}
\omega \sum_{t=0}^\infty \left( ||r^{t+1}||^2 + ||z^{t+1}-z^t||^2   \right) \leq V^0,
\end{equation}
which implies $r^t \rightarrow 0$ and $z^t  \rightarrow z^{t+1}$ as $t \rightarrow \infty$.

Now, we define $p^t = f(x^t) + g(z^t)$ and show $p^t$ converges.
Because $(x^*, z^*, y^*)$ is a saddle point,
\begin{equation}
\mathcal{L} (x^*, z^*, y^*) \leq \mathcal{L} (x^{t+1}, z^{t+1}, y^*).
\end{equation}
That is, using $x^* = z^*$ at the saddle point,
\begin{equation}
p^* \leq p^{t+1} + y^{*T} r^{t+1} + (\omega /2) ||r^{t+1} ||^2,
\end{equation}
which implies that upper bound of $p^* - p^{t+1} $ is
\begin{equation}
p^* - p^{t+1} \leq y^{*T} r^{t+1}.
\end{equation}
Lower bound of $p^* - p^{t+1} $ (proof in \cite{ykim16trip}) is
\begin{equation}
p^* - p^{t+1}\geq (y^{t+1})^T r^{t+1} + \omega (z^{t+1} - z^t)^T (r^{t+1} + (z^{t+1}-z^*)).
\label{eq:lowerbound}
\end{equation}
The upper and lower bounds go to zero because $r^t \rightarrow 0$ and $z^t  \rightarrow z^{t+1}$ as $t \rightarrow \infty$, i.e., $\lim_{t \rightarrow \infty} p^t = p^*$. Thus, the objective function $\Psi$ of our federated optimization converges.

\subsection{Privacy analysis}
In our \emph{Honest-but-Curious} adversary scenario, privacy of patient data is preserved because patient-level data are not disclosed to the both server and hospitals.
The server and hospitals cannot access to unintended fine-grained local information. The local data are only accessible to the corresponding hospital. 
The server also cannot indirectly learn patient data from the local factor matrices.
After receiving $\mathbf{A}_k^{(n)}$, the server might try to do reverse-engineering through Eq. (\ref{eq:Ain}).
However, server cannot access to $\Pi_k^{(n)}$ because $\mathbf{A}_k^{(1)}$ from patient mode is not shared.
If server accesses to $\Pi_k^{(n)}$ by any chance as $\mathbf{A}_k^{(1)}$ is leaked, 
reverse-engineering cannot still restore patient-level data.  
That is, the matricized unknown observed tensor (patient data) has an equation in form of $\mathbf{A}_k^{(n)}=\mathbf{O}_{(n)k} \Pi_k$ after removing all the known values in Eq. (\ref{eq:Ain}) for simplicity.
The size of the unknown information in $\mathbf{O}_{(n)k} $ is $I_n \times (I_{1k} \cdots I_{n-1} I_{n+1} \cdots I_N)$, and the size of  $\Pi_k^{(n)}$ and $\mathbf{A}_k^{(n)}$ is $(I_{1k} \cdots I_{n-1} I_{n+1} \cdots I_N) \times r$ and $I_n \times r$, respectively.
Element-wise computation generates only $I_n \cdot r$ equations for the unknown $I_1 \cdots  I_N$ values.
Server cannot recover the unknown values from the $I_n \cdot r$ equations that is less than the number of unknown values ($r$ is always selected as $I_n \cdot r  \ll I_1 \cdots  I_N$).

Hospitals also cannot learn other hospitals' data from the global factor matrix.
If hospital $k'$ knows all the information
of Eq. (\ref{eq:Aw}) for global factor matrix by any chance,
hospital $k'$ cannot restore other hospitals' local factor matrix $\mathbf{A}_k^{(n)}$. 
If the hospital $k'$ can access to $\mathbf{A}_k^{(n)}$ by any chance,
$\mathbf{A}_k^{(n)}$ has still insufficient information to recover the data as shown in the case of server. 

\subsection{Secure alignment of feature modes}
\begin{figure}[t]
\centering
\includegraphics[width=0.5\textwidth]{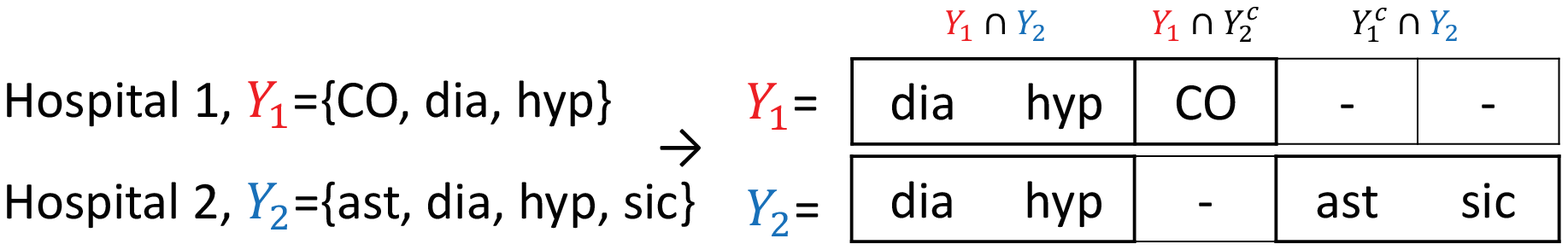}
\vspace{-5mm}
\caption{Example of secure alignment on feature mode.}
\label{fig:eg}
\end{figure}

In Section \ref{sec:formulation}, we first assume that hospitals have the same element set for each feature in feature modes, but in practice, hospitals may have different elements.
For example, Hospital 1 and Hospital 2 have set of diagnosis: $Y_1$ and $Y_2$ (Fig. \ref{fig:eg}),
but each index of $Y_1$ and $Y_2$ refers to a different element.
In this case, before concatenating the local tensors, the index of feature modes should refer to the same element among hospitals.
Thus, we introduce a secure alignment method for feature modes by which hospitals do not reveal the elements they have and get an integrated and sorted view on the elements of feature modes.
This secure alignment enables hospitals to have only the position of its elements without knowing other hospitals' elements as like $Y_1$ and $Y_2$ are aligned to make the index from two sets refer to the same element (Fig. \ref{fig:eg}).
For each feature mode, hospital $k$ assigns integer values to the set of elements $Y_k$ (eg. ICD9 codes). 
Hospitals use polynomial properties of set intersection \cite{kissner2005privacy}:
\begin{lemma}
A polynomial function of $y$ that represents set of elements $Y_k=\{y_{ik}\}_{i=1}^{I} $ at hospital $k$ is 
$f_k (y)=(y-y_{1k})(y-y_{2k}) \cdots (y-y_{Ik}) = \sum_{i=0}^I a_i y^i.$
A $y_{ik}$ is an element of $Y_k$ ($y_{ik} \in Y_k$) if and only if $f_k(y_{ik})=0$.
\label{lemma1}
\end{lemma}
\begin{lemma}
A polynomial function that represents intersection of $Y_k$ and $Y_{k'}$ ($Y_k \cap Y_{k'}$) is $f_k*r + f_{k'}*s$ where $r, s$ are polynomial functions with $gcd(r,s)=1$.
Given $f_k*r + f_{k'}*s$, one cannot learn individual elements on $Y_1$ and $Y_2$ other than elements in $Y_1 \cap Y_2$.
\label{lemma2}
\end{lemma}
Hospitals express $Y_k$ as a polynomial function (or in short polynomial) $f_k$ by Lemma \ref{lemma1}.
To prevent the factorization of the polynomial, hospital $k$ multiplies a term $r=(y-\alpha)$ to $f_k$ (=$f_k*r$), where $\alpha$ is a random prime number that is selected with overwhelming probability that the $\alpha$ does not represent any element from $Y_k$. For simplicity, $f_k*r$ is denoted as $f_k$. Because computing the polynomials with large $|Y_k|$ can cause computational overhead, hospitals compute the polynomials' modulus (denoted as \%) by a random prime number $P$ ($P > y_k$) instead of the polynomial itself, i.e., hospitals compute
$f_k \% P = \left[\sum_{i=0}^I (a_i \%P)y^i \right] \% P$
by equivalence of modulus operation and use it instead of $f_k$.

Then server receives $f_k \% P$ from hospitals.
To find a pairwise intersection between hospital $k$ and $k'$, server computes a pairwise sum of polynomials as
$\left[f_k \%P + f_{k'}\%P \right] \%P= \left[ f_k  + f_{k'} \right] \%P$,
which refers to the polynomial for intersection between hospital $k$ and $k'$. Server repeats this procedure for every pair of $k$ and $k'$ ($k' \neq k$), and send the $K-1$ polynomials to each hospital. Hospital $k$ then checks whether its element $y_k \in Y_k$ is in the pairwise intersection of hospital $k$ and other hospital $k'$, that is, if $[f_k(y_k) + f_{k'}(y_k)] \%P = 0$, then $y_k \in Y_k \cap Y_{k'}$ by Lemma \ref{lemma2}.


By combining all the pairwise intersection with $K-1$ hospitals, hospital $k$ checks whether the element $y_k$ is in the intersection of all the $K$ hospitals. For example, combining the pairwise intersection of $f_1(y_1)+f_2(y_1)=0$ (i.e., $y_1 \in  Y_1 \cap Y_2$) and $f_1(y_1)+f_3(y_1) \neq 0$ (i.e., $y_1 \notin  Y_1 \cap Y_3$) gives $y_1 \in Y_1 \cap Y_2 \cap Y_3^{c}$. After obtaining $2^{K-1}$ intersections with $Y_k$, hospital $k$ sends the size of $2^{K-1}$ intersections to server.
Server collects the size of intersection from all the $K$ hospitals and obtain the size of all the $2^K-1$ combinations of intersections (the number of combinations is two cases, whether in or out, for every $Y_k$ except one case when $Y_1^c \cap \ldots \cap Y_K^c$). 
Finally, hospitals receive the size of $2^K-1$ intersection, and align their elements according to the size information. 
Hospitals have the same order of these intersections such as $Y_1 \cap Y_2, Y_1 \cap Y_2^c, Y_1^c \cap Y_2$ (Fig. \ref{fig:eg}).
The elements within the intersections are sorted. Thus, all hospitals have the same size and order of elements for every feature mode.

\section{Experiments}
We evaluate \textsc{Trip} by measuring computational performance (accuracy and time) and deriving meaningful phenotypes. 
We compare \textsc{Trip} with two baselines: \\
\textbf{i) Central model: }Ordinary tensor factorization method for phenotyping. Regardless of privacy problem that concerns data sharing, this model runs on a central server where all the patient data are combined \cite{wang2015rubik}. \\
\textbf{ii) Local model: }We devise an intuitive local model, by which hospitals run the central model at their sides and send the final factor matrices of feature modes to server. Server averages the factor matrices and sends the averaged factor matrices back to hospitals without iterative updating like \textsc{Trip}.
Because each column in factor matrices can represent different phenotypes over hospitals, before averaging the matrices, server sorts the columns of each hospital's factor matrix so that all hospitals have the same phenotypes at each column. For all feature modes $n$, server first chooses a pivot hospital $k_p$ and 
computes cosine similarity between every pair of $r_p$th and $r$th column from factor matrix of hospital $k_p$ and other hospitals $k$ as
$\text{similarity}= \cos (\mathbf{A}_{k_p}^{(n)}(:, r_p), \mathbf{A}_{k}^{(n)}(:, r))$
where $\forall k \neq k_p, ~ \forall  r \neq r_p $.
Server then finds the most similar combination of $r_p$ and $r$ for all pairs. Finding the best combination that matches multiple items (columns) to multiple items can be solved in polynomial time by Hungarian method \cite{kuhn1955hungarian}. 
Finally, server changes the order of columns in $\mathbf{A}_{k}^{(n)}$ according to the combination so that each column from  $\mathbf{A}_{k_p}^{(n)}$  and $\mathbf{A}_{k}^{(n)}$  refer to the same phenotype. 


\begin{figure*}[t]
\centering
\subfloat[RMSE vs \# nonzeros ]{
\includegraphics[width=0.25\textwidth]{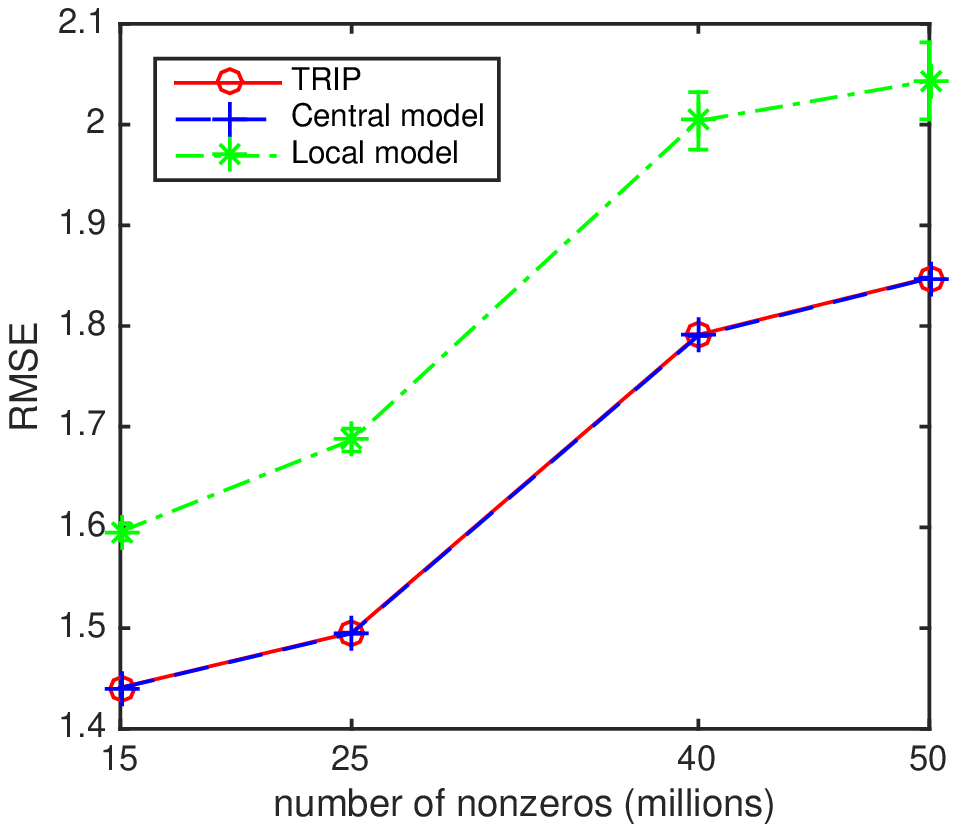}
\label{fig:rmse_nonzeros}
}
\subfloat[Time vs. \# nonzeros]{
\includegraphics[width=0.25\textwidth]{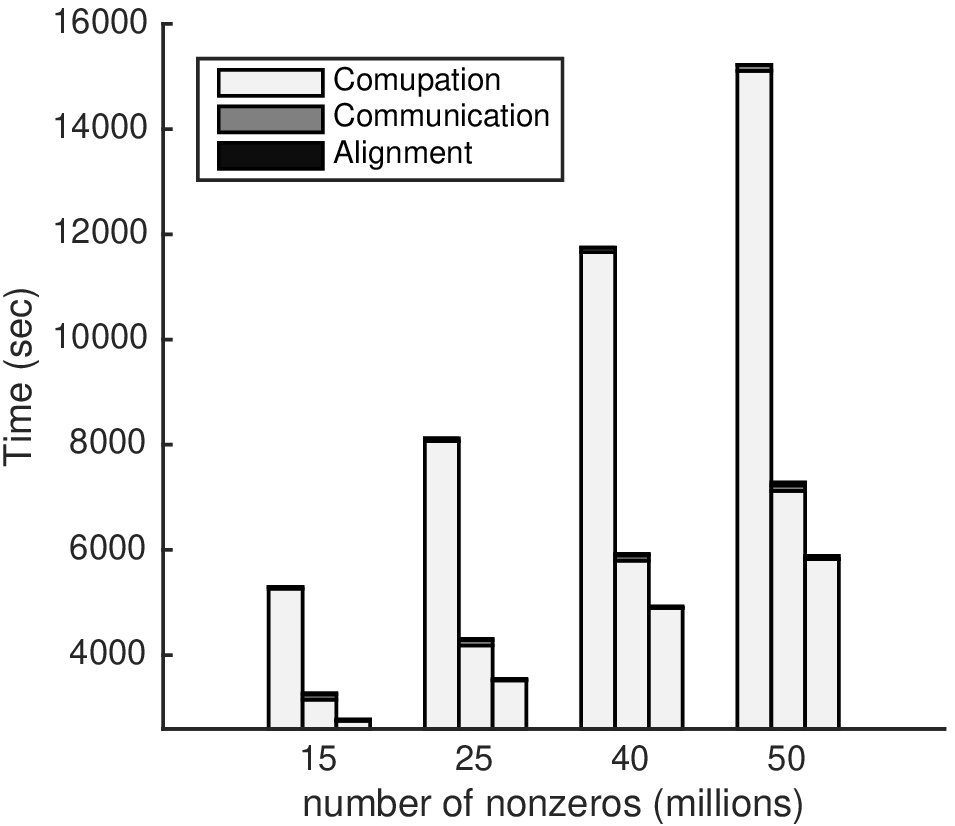}
\label{fig:time_nonzero}
}
\subfloat[MIMIC-III 15M]{
\includegraphics[width=0.25\textwidth]{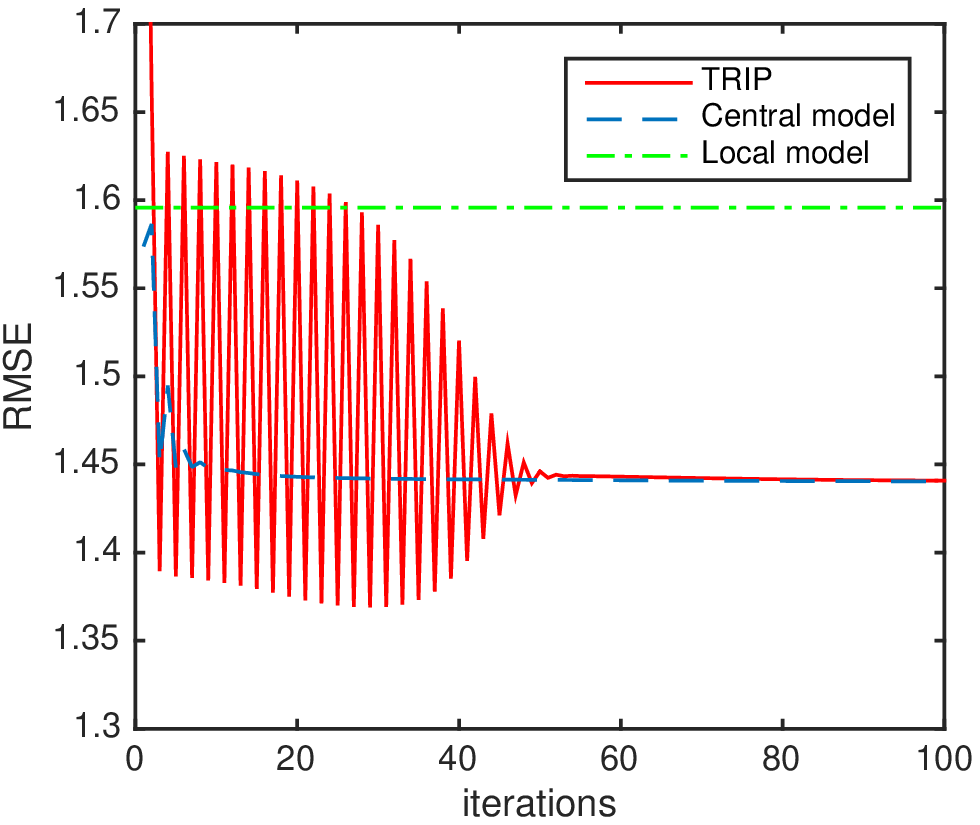}
\label{fig:rmse_nonzero_15}
}
\subfloat[MIMIC-III 50M]{
\includegraphics[width=0.25\textwidth]{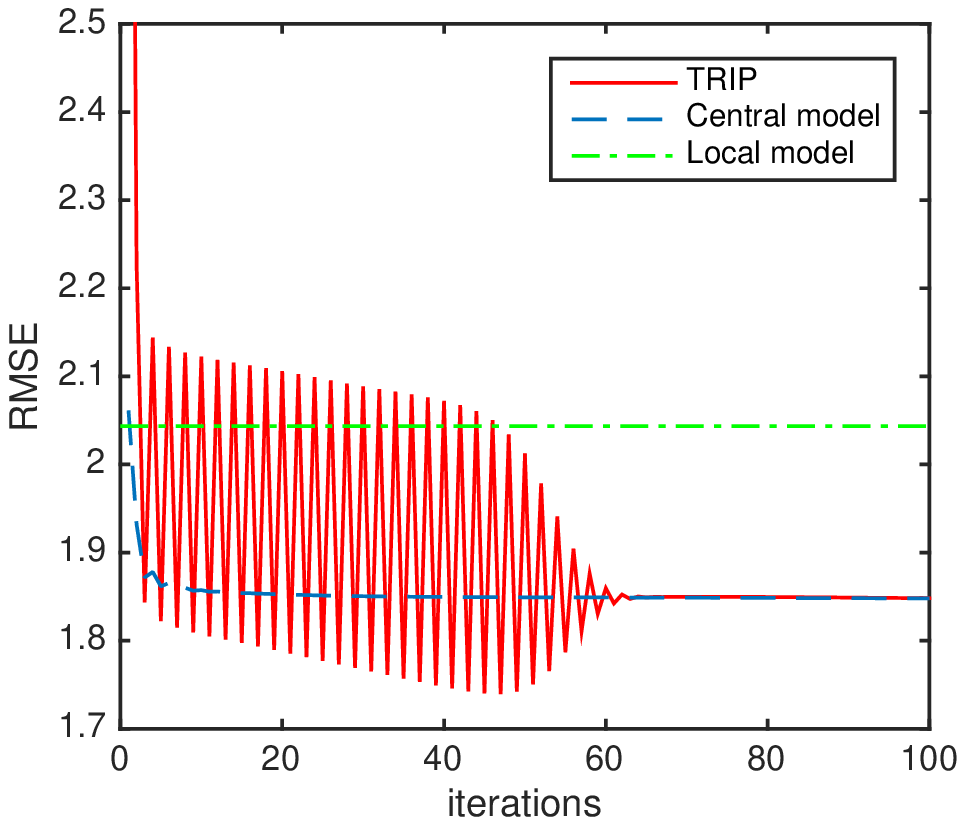}
\label{fig:rmse_nonzero_50}
}
\vspace{-2mm}
\caption{RMSE and total time over the number of nonzeros (Fig. \ref{fig:rmse_nonzeros}, \ref{fig:time_nonzero}). The first, second, and third stacked bars in Fig. \ref{fig:time_nonzero} refer to central model, \textsc{Trip}, and local model, respectively. RMSE of \textsc{Trip}, central model, and local model over iteration (Fig. \ref{fig:rmse_nonzero_15}, \ref{fig:rmse_nonzero_50}).}
\label{fig:iter_nonzero}
\vspace{-3mm}
\end{figure*}
\begin{figure*}[h]
\centering
\subfloat[RMSE vs. \# hospitals]{
\includegraphics[width=0.25\textwidth]{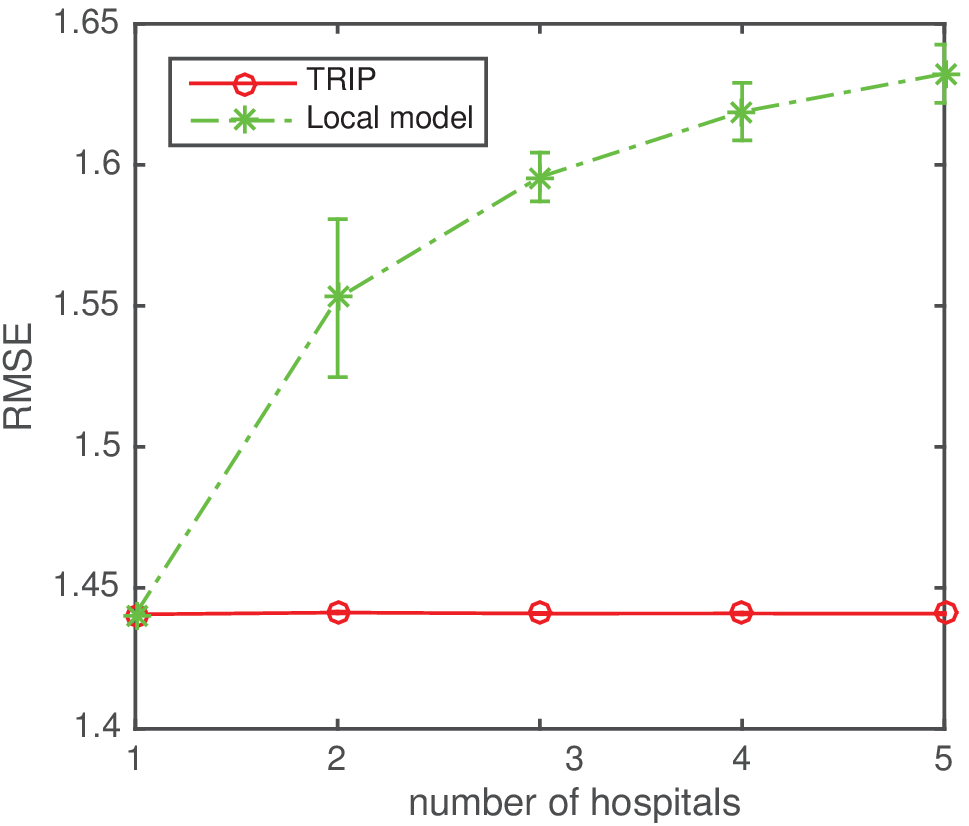}
\label{fig:rmse_hospitals}
}
\subfloat[RMSE vs. skewness]{
\includegraphics[width=0.25\textwidth]{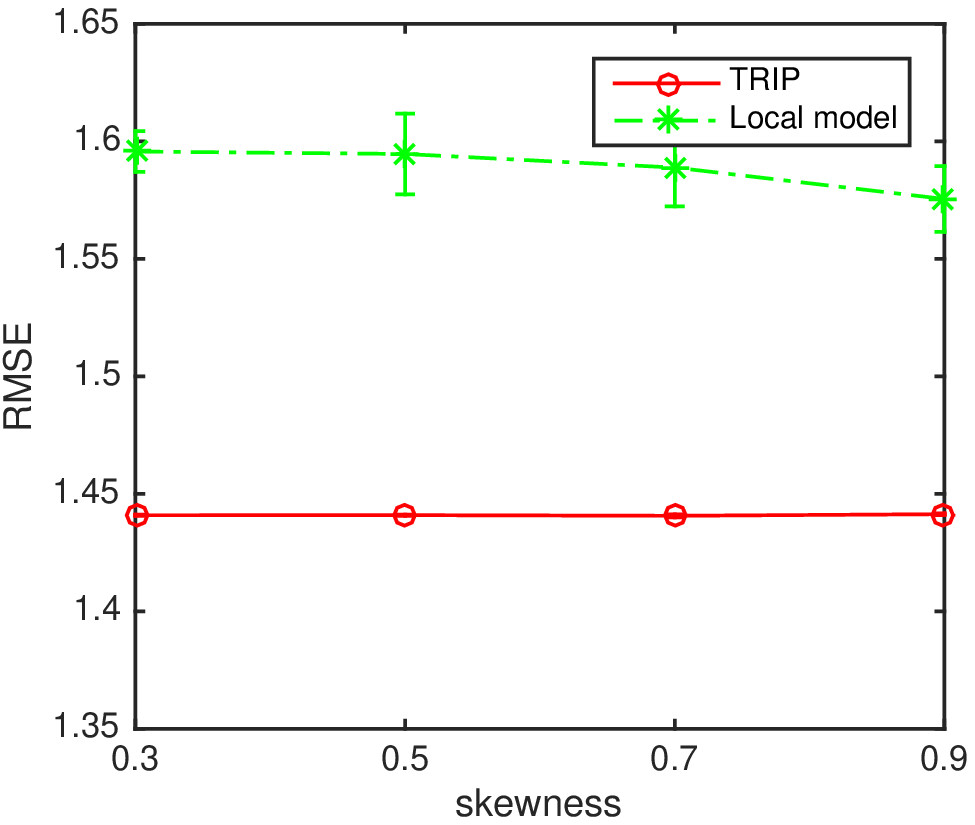}
\label{fig:rmse_skewness}
}
\subfloat[Time vs. \# hospitals]{
\includegraphics[width=0.25\textwidth]{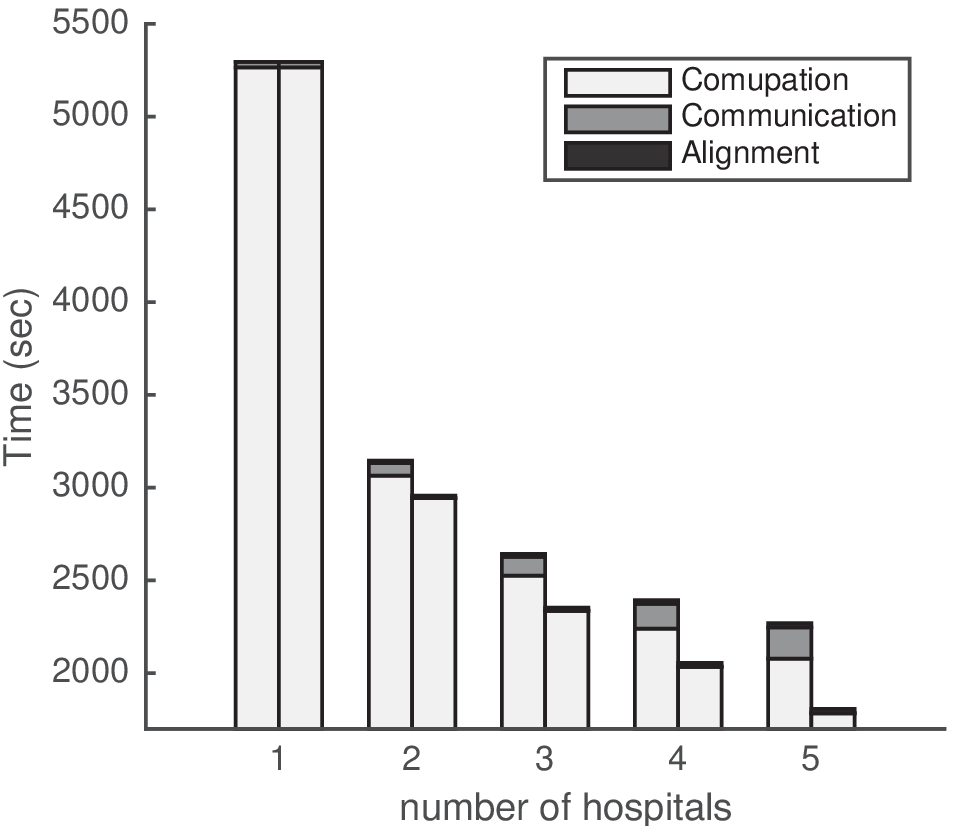}
\label{fig:time_hospitals}
}
\subfloat[Time vs. skewness ]{
\includegraphics[width=0.25\textwidth]{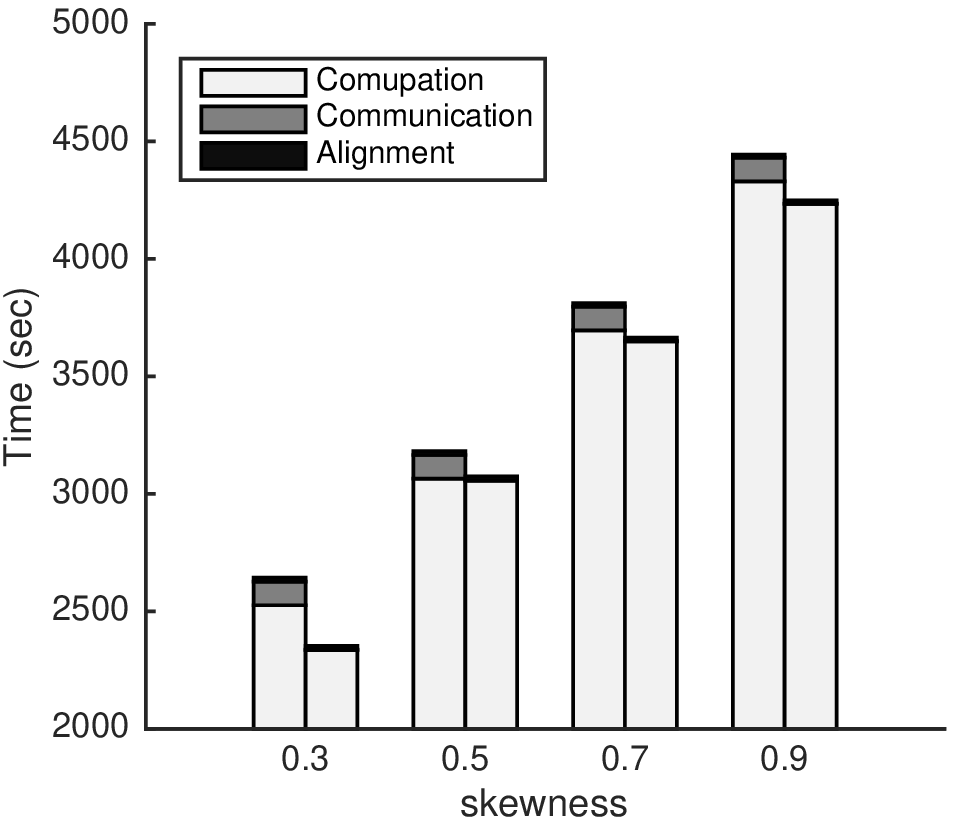}
\label{fig:time_skewness}
}
\vspace{-2mm}
\caption{RMSE over the number of hospitals (Fig. \ref{fig:rmse_hospitals}) and skewness (Fig. \ref{fig:rmse_skewness}). Total time over the number of hospitals (Fig. \ref{fig:time_hospitals}) and skewness (Fig. \ref{fig:time_skewness}). The former and latter stacked bars in Fig. \ref{fig:time_hospitals}, \ref{fig:time_skewness} refer to \textsc{Trip} and local model, respectively.}
\label{fig:robustness}
\end{figure*}
\subsection{Accuracy and Time}
We use a large publicly available dataset MIMIC-III containing de-identified health-related data associated with over forty thousand patients who stayed in critical care units of the Beth Israel Deaconess Medical Center between 2001 and 2012 \cite{MIMICII}. MIMIC-III includes information such as demographics, laboratory test results, procedures, medications, caregiver notes, imaging reports, and mortality.
We construct a 3-order tensor with patients, laboratory test results, and medication. 
The tensor value is the number of co-occurrences of abnormal lab results and medication from the same patient within specific time window. We generate four datasets as setting the time window as 3 hours, 6 hours, 1 day, or 7 days, and have the number of nonzero values of around 15 million (M), 25M, 40M, and 50M, respectively. The size for 7-day-window tensor (MIMIC-III 50M) is 38,035 patients by 3,229 medications by 304 lab results. Because duplicated co-occurrence can be counted with large time window, we set the maximum value of count as three, which is a median of 1-day-window tensor (MIMIC-III 40M). The count value larger than three is truncated to three. 

We evaluate accuracy and time of \textsc{Trip} compared to two baseline models by varying the number of nonzero values, hospitals, and skewness (for unevenly distributed patients).
We measure accuracy using root mean square error (RMSE) between the factorized tensor and the observed tensor.
We also measure time elapsed by adding time for computation, communication, and alignment.
Because \textsc{Trip} and local model distribute computation of local tensors to hospitals,
we consider the computation time on the local tensors as the largest computation time on one hospital. 
The communication time is measured as the total number of communicated bytes between server and hospitals divided by data transfer rate of 15 MB/sec. 
The communication time for central model is time for transferring the local patient data to server.  
We repeat the evaluation ten times and average them.
We run the models until it converges before maximum iteration 100.
The rank is set to ten. $\lambda$ is set to $10^{-2}$ after trying $10^{-3}, 10^{-2}, 10^{-1}, 1$, and 10. 

\subsubsection{Number of nonzeros}
We use the four MIMIC-III datasets that have 15M, 25M, 40M, and 50M nonzero values. We assume that MIMIC-III datasets are distributed in three hospitals, on which the patients are randomly distributed as the same size. 
As a result, \textsc{Trip} has low RMSE as much as central model and resembles central model for all the four datasets (Fig. \ref{fig:rmse_nonzeros}). 
For MIMIC-III 15M, RMSE from central model converges to 1.4404. Similarly, the RMSE from \textsc{Trip} starts to stable at around 50 iterations and converges to 1.4409 (Fig. \ref{fig:rmse_nonzero_15}). 
Both of the RMSE from central model and \textsc{Trip} are significantly smaller than that of local model (1.5957).  
MIMIC-III 50M dataset also shows similar convergence. RMSE from \textsc{Trip} starts to stable at around 60 iterations and converges to 1.8482, which is also similar to RMSE from central model, 1.8479 (Fig. \ref{fig:rmse_nonzero_50}).
MIMIC-III 25M shows the RMSE of 1.4955 from \textsc{Trip}, 1.4947 from central model, 1.6867 from local model.
MIMIC-III 40M shows the RMSE of 1.7913 from \textsc{Trip}, 1.7903 from central model, 2.0037 from local model. 
Convergence results on MIMIC-III 25M and 40M can be found in our technical report \cite{ykim16trip}.

In addition, total time elapsed for \textsc{Trip} (and local model) is much faster (half less) than central model in all datasets (Fig. \ref{fig:time_nonzero}). 
\textsc{Trip} reduces computation time by distributing updating procedures to de-centralized hospitals; 
consequently, \textsc{Trip} reduces total time elapsed although sacrificing communication and alignment time.
For the datasets of 15M, 25M, 40M, and 50M, 
computation time from \textsc{Trip} is 3,152, 4,183, 5,796, and 7,125 seconds,
and computation time from central model is 5,266, 8,068, 11,661, and 15,105 seconds, which take majority of total time. 
Communication time from \textsc{Trip} is around 100.7 seconds for all the cases, and
communication time from central model is 30.1, 53.9, 85.0, and 114.4 seconds.
Note that \textsc{Trip} saves not only computation time but also communication time with large dataset (MIMIC-III 50M). 
Alignment time for \textsc{Trip} takes 22.6, 26.1, 29.3, and 59.0 seconds, which is negligible compared to computation time.
Based on those observations, we can see that \textsc{Trip} efficiently resembles central model without no cost for privacy even at reduced time owing to distributed computation.

\subsubsection{Number of hospitals}
Using MIMIC-III 15M dataset, we partition the patients evenly into one to five hospitals. RMSE with one hospital refers to RMSE of central model. 
We observed that RMSE of \textsc{Trip} is stable as the number of hospitals increases, and is similar to RMSE of central model 1.4404, whereas RMSE of local model increase (Fig. \ref{fig:rmse_hospitals}) with large variance.
That is, compared to local model, in which local factorized tensors are diverged, \textsc{Trip} is robust on the finely split data.  
It means that phenotypes from \textsc{Trip} are accurate and not biased even with many small sized patient data. 
 

Total time of \textsc{Trip} and local model are significantly faster than that of central model. 
As the number of hospitals increases and the patient data are spread more, the total time of \textsc{Trip} and local model decrease (Fig. \ref{fig:time_hospitals}).
Specifically, 
computation time for \textsc{Trip} and local model decrease because more hospitals distribute the computation, and
communication time for \textsc{Trip} slightly increases, whereas communication time for local model is negligibly short.
Alignment time is negligible for both \textsc{Trip} and local model.

\begin{table*}[t]
\small
\centering
\caption{Phenotypes from \textsc{Trip}, central model on UCSD1+UCSD2, \textcolor{blue}{UCSD1}, and \textcolor{red}{UCSD2}. Some phenotypes appear in \textcolor[HTML]{9600FF}{both UCSD hospitals}}
\label{table:phenotpyes}
\vspace{-3mm}
\begin{tabular}{c | L{3.8cm} L{3.5cm} L{3.8cm} L{3.8cm} l }
\hline
Rank & \multicolumn{1}{c}{TRIP}                                                     & \multicolumn{1}{c}{  UCSD1+UCSD2}                                            & \multicolumn{1}{c}{UCSD1}                                                & \multicolumn{1}{c}{UCSD2}                                       \\ \hline
1    & {\color[HTML]{0013FF} Coronary artery disease with diabetes \& hypertension} & {\color[HTML]{0013FF} Diabetic with hypertension}                            & {\color[HTML]{0013FF} Diabetic with hypertension}                            & {\color[HTML]{FF0000} Cystic fibrosis with pancreatic involvement} \\ \hline
2    & {\color[HTML]{0013FF} Diabetic with hypertension}                            & {\color[HTML]{FF0000}Cystic fibrosis with pancreatic involvement}                                 & {\color[HTML]{0013FF} Coronary artery disease with diabetes \& hypertension} & {\color[HTML]{FF0000} Cystic fibrosis with pulmonary exacerbation} \\ \hline
3    & {\color[HTML]{0013FF} Chronic obstructive pulmonary disease (COPD) exacerbation}                                     & {\color[HTML]{0013FF} Coronary artery disease with diabetes \& hypertension} & {\color[HTML]{0013FF} COPD exacerbation}                                     & {\color[HTML]{FF0000} Neurogenic bladder with abdominal pain}      \\ \hline
4    & {\color[HTML]{FF0000} Constipation}                                          & {\color[HTML]{0013FF} Hypertension}                                          & {\color[HTML]{0013FF} Decompensated cirrhosis}                               & {\color[HTML]{9600FF} Non-specific gastrointestinal complaints}                  \\ \hline
5    & {\color[HTML]{FF0000} Cystic fibrosis with pancreatic involvement}           & {\color[HTML]{0013FF} COPD exacerbation}                                     & {\color[HTML]{9600FF} Non-specific gastrointestinal complaints}                            & Diabetes                                                           \\ \hline
6    & {\color[HTML]{0013FF} Decompensated cirrhosis}                               & {\color[HTML]{FF0000} Constipation}                                          & Non-specific complaints                                                      & {\color[HTML]{FF0000} Constipation}                                \\ \hline
7    & {\color[HTML]{9600FF} Non-specific gastrointestinal complaints}                            & {\color[HTML]{0013FF} Decompensated cirrhosis}                               & COPD w/o exacerbation                                                        & Anxiety with gastrointestinal complaints                                         \\ \hline
8    & {\color[HTML]{FF0000} Cystic fibrosis with pulmonary exacerbation}           & Non-specific complaints                                                      & Acute on chronic pain                                                        & Cystic fibrosis with pneumonia                                     \\ \hline
9    & Sickle cell/chronic pain crisis                                              & Sickle cell/chronic pain crisis                                              & COPD with Pneumonia                                                          & Non-specific complaints                                            \\ \hline
10   & Non-specific complaints                                                      & {\color[HTML]{FF0000} Neurogenic bladder with abdominal pain}                & Anxiety with hypertension                                                    & Lymphoma                                                           \\ \hline
\end{tabular}
\end{table*}

\subsubsection{Skewness}
We partition the patients in MIMIC-III 15M unevenly in three hospitals. One hospital takes 1/3 (evenly distributed), 0.5, 0.7, and 0.9 of patients, and the other two hospitals take the remaining patients evenly. 
Note that elements in feature mode are still overlapped enough among hospitals. 
We observed that RMSE of \textsc{Trip} is stable although patients are distributed unevenly, whereas RMSE of local model is higher than that of \textsc{Trip} with large variance. 
Factorized tensor of local model can be inaccurate because the local factorized tensor from a small sized hospital can be biased and far different from others' results. 
However, the hospital can benefit from \textsc{Trip} by overcoming this bias and producing a generalized results.


Total time of \textsc{Trip} and local model increase (Fig. \ref{fig:time_skewness}) as the skewness increases. 
Time for computation increases because computational overhead occurs on one hospital with large data, and time for communication and alignment does not increase significantly.


\subsection{Phenotype discovery}

We use de-identified EHRs dataset from University of California, San Diego (UCSD) Medical Center with 8,022 patients by 748 medications by 299 diagnoses.
Specifically, it is from two hospitals that have 4,703 patients (UCSD1) and 3,319 patients (UCSD2). We construct a 3-order tensor with patient, medication, and diagnosis mode with around 1.6 million of non-zero elements. The value of tensor is the number of co-occurrences of medication and diagnosis event from the same patient at the same visit. 

We discover phenotypes from \textsc{Trip} and compare them with phenotypes from central model and individual central model of
two hospitals in UCSD (i.e., run central model independently at UCSD1 and UCSD2). 
$\lambda$ is set 1 to derive more distinct phenotypes than those from MIMIC-III. 
A domain expert summarizes the factorized tensor into clinically meaningful phenotypes. The phenotypes consist of set of diagnoses and its corresponding medications. Due to limited space, medication factors in phenotypes are omitted and can be found in our project website \cite{ykim16trip}. 

As a result, \textsc{Trip} discovers unbiased and hidden phenotypes compared to the phenotypes from two individual central models (UCSD1, UCSD2).
The phenotypes from \textsc{Trip} contain top-ranked phenotypes from UCSD1 and UCSD2, and are similar to phenotypes from combined central model, UCSD1+UCSD2 (Table \ref{table:phenotpyes}). 
The phenotypes from \textsc{Trip} consist of top five phenotypes from UCSD1 and top four from UCSD2.
The phenotypes from \textsc{Trip} are also the same with phenotypes from central model except gastrointestinal complaints and neurogenic bladder. 
Without our federated model, the two individual hospitals could derive biased phenotypes that are only fitted to the local data. 
It means that \textsc{Trip} can effectively resemble central model without cost for privacy.

In addition, \textsc{Trip} discovers a new phenotype, \emph{sickle cell/chronic pain crisis}, that is contained in neither of UCSD1 and UCSD2. 
This phenotype consists of diagnoses related to \emph{sickle cell diseases} or \emph{chronic pain crisis} and corresponding medications (Table \ref{table:ex_phenotype}). 
Based on physician's judgement, this phenotype is clinically meaningful in that sickle cell disease usually accompanies chronic pain such as constipation, back/neck pain, headache, (pruritic disorder, insomnia, and wheezing.
\emph{sickle cell/chronic pain crisis} is not dominant in individual hospital but is dominant in overall perspective.
Note that RMSE of \textsc{Trip} is low as much as RMSE of central model while reducing total time (Table \ref{table:time_ucsd}), and RMSEs of two individual UCSD datasets are lower than others because those two use separated small local datasets. 
Also, note that communication time of the central model is due to transferring the data. 


%

\begin{table}[t]
\centering
\caption{Detailed phenotype of \emph{sickle cell/chronic pain crisis} 
}
\label{table:ex_phenotype}
\begin{tabular}{L{8.5cm}}
\hline
Diagnosis \\ \hline
Sickle cell disease NOS,                           
Hb-SS disease with crisis,    
Constipation NOS,
Pruritic disorder NOS,
Generalized pain,
Headache,
Insomnia,  
Chronic pain syndrome,
Wheezing    
\\ \hhline{=}
Medication
\\ \hline
Hydroxyurea,
Deferasirox, Docusate, Diphenhydramine, Hydromorphone, Acetaminophen, Zolpidem, Folic Acid, Baclofen 
\\ \hline
\end{tabular}
\end{table}

\begin{table}[t]
\centering
\caption{RMSE and time [sec] elapsed. Central model is run on UCSD1,2, UCSD1, UCSD2.
}
\label{table:time_ucsd}
\begin{tabular}{c|cccc}
\hline
              & Trip    & UCSD1,2 & UCSD1 & UCSD2 \\ \hline
RMSE          & 1.2304  & 1.2327        & 1.2267    & 1.1778   \\
Computation   & 446.5   & 656.9         & 432.8     & 314.6    \\
Communication & 15.1196 & 3.4533        & 0         & 0        \\
Alignment     & 1.2052  & 0             & 0         & 0        \\ \hline
\end{tabular}
\vspace{-2mm}
\end{table}

\section{Conclusions}
This paper presents \textsc{Trip}, a federated tensor factorization for computational phenotyping without sharing patient-level data. We developed secure data harmonization and privacy-preserving computation procedures based on ADMM, and analyzed that \textsc{Trip} ensure the confidentiality of patient-level data. Experimental results on data from MIMIC-III and UCSD medical center demonstrated that our framework resembles the central model very well. \textsc{Trip} is also accurate even with small or skewly distributed patient data, and fast compared to the central model. 
We also showed that \textsc{Trip} discovers phenotypes as the central model with combined patient data does, which are unbiased or not discovered (hidden) phenotypes from each hospital.  
As a result, \textsc{Trip} can help derive useful phenotypes from EHR data to overcome policy barriers due to the privacy concerns. We plan to apply it to much larger scale datasets in the future and facilitate the discovery of novel and important ``phenotypes'' to support clinical research and precision medicine.
\\
\textbf{Acknowledgments. }
\small
We thank Dr. Robert El-Kareh, MD from University of California San Diego to annotate the computed phenotype.
This research was partly supported by NHGRI (R00HG008175, R01HG008802), NIH (R01GM-118609, R01GM-118574), NLM (R00LM011392, R21LM012060), NIPA (NIPA-2014-H0201-14-1001), NRF (2012M3C4A7033344), MSIP/IITP (B0101-15-0307), and MOTIE (10049079). 

\normalsize

%
\bibliographystyle{abbrv}
\bibliography{ref,xj,jimengref,tensor}  

\begin{thebibliography}{10}

\bibitem{Bennett:2012gc}
C.~Bennett, T.~Doub, and R.~Selove.
\newblock {EHRs connect research and practice: Where predictive modeling,
  artificial intelligence, and clinical decision support intersect}.
\newblock {\em Health Policy and Technology}, 1(2):105--114, June 2012.

\bibitem{boyd2011distributed}
S.~Boyd, N.~Parikh, E.~Chu, B.~Peleato, and J.~Eckstein.
\newblock Distributed optimization and statistical learning via the alternating
  direction method of multipliers.
\newblock {\em Foundations and Trends{\textregistered} in Machine Learning},
  3(1):1--122, 2011.

\bibitem{carroll1970analysis}
J.~D. Carroll and J.-J. Chang.
\newblock Analysis of individual differences in multidimensional scaling via an
  n-way generalization of ''eckart-young'' decomposition.
\newblock {\em Psychometrika}, 35(3):283--319, 1970.

\bibitem{Carroll:2011ue}
R.~J. Carroll, A.~E. Eyler, and J.~C. Denny.
\newblock {Na{\"\i}ve electronic health record phenotype identification for
  rheumatoid arthritis}.
\newblock {\em Proceedings of the American Medical Informatics Association
  Annual Symposium}, pages 189--196, Jan. 2011.

\bibitem{Carroll:2012jr}
R.~J. Carroll, W.~K. Thompson, A.~E. Eyler, A.~M. Mandelin, T.~Cai, R.~M. Zink,
  J.~A. Pacheco, C.~S. Boomershine, T.~A. Lasko, H.~Xu, E.~W. Karlson, R.~G.
  Perez, V.~S. Gainer, S.~N. Murphy, E.~M. Ruderman, R.~M. Pope, R.~M. Plenge,
  A.~N. Kho, K.~P. Liao, and J.~C. Denny.
\newblock {Portability of an algorithm to identify rheumatoid arthritis in
  electronic health records.}
\newblock {\em Journal of the American Medical Informatics Association},
  19(1e):e162--e169, June 2012.

\bibitem{CMS11}
K.~Chaudhuri, C.~Monteleoni, and A.~Sarwate.
\newblock {Differentially Private Empirical Risk Minimization}.
\newblock {\em Journal of Machine Learning Research (JMLR)}, 12:1069--1109,
  July 2011.

\bibitem{choi2014dfacto}
J.~H. Choi and S.~Vishwanathan.
\newblock Dfacto: Distributed factorization of tensors.
\newblock In {\em Advances in Neural Information Processing Systems}, pages
  1296--1304, 2014.

\bibitem{Cimino:2000uh}
J.~J. Cimino.
\newblock {From data to knowledge through concept-oriented terminologies:
  Experience with the Medical Entities Dictionary}.
\newblock {\em Journal of the American Medical Informatics Association},
  7(3):288--297, May 2000.

\bibitem{de2014distributed}
A.~L. De~Almeida and A.~Y. Kibangou.
\newblock Distributed large-scale tensor decomposition.
\newblock In {\em Acoustics, Speech and Signal Processing (ICASSP), 2014 IEEE
  International Conference on}, pages 26--30. IEEE, 2014.

\bibitem{dwork06}
C.~Dwork.
\newblock {Differential privacy}.
\newblock {\em International Colloquium on Automata, Languages and
  Programming}, 4052(d):1--12, 2006.

\bibitem{ebadollahi_predicting_2010}
S.~Ebadollahi, J.~Sun, D.~Gotz, J.~Hu, D.~Sow, and C.~Neti.
\newblock Predicting patient's trajectory of physiological data using temporal
  trends in similar patients: A system for near-term prognostics.
\newblock {\em {AMIA} Annual Symposium Proceedings}, 2010:192--196, 2010.
\newblock {PMID:} 21346967.

\bibitem{Greengard:2013iu}
S.~Greengard.
\newblock {A new model for healthcare}.
\newblock {\em Communications of the ACM}, 56(2):17--19, 2013.

\bibitem{ho2014limestone}
J.~C. Ho, J.~Ghosh, S.~R. Steinhubl, W.~F. Stewart, J.~C. Denny, B.~A. Malin,
  and J.~Sun.
\newblock Limestone: High-throughput candidate phenotype generation via tensor
  factorization.
\newblock {\em Journal of biomedical informatics}, 52:199--211, 2014.

\bibitem{ho_marble:_2014}
J.~C. Ho, J.~Ghosh, and J.~Sun.
\newblock Marble: {High}-throughput {Phenotyping} from {Electronic} {Health}
  {Records} via {Sparse} {Nonnegative} {Tensor} {Factorization}.
\newblock In {\em KDD14},
  pages 115--124, New York, NY, USA, 2014. ACM.

\bibitem{hong2016convergence}
M.~Hong, Z.-Q. Luo, and M.~Razaviyayn.
\newblock Convergence analysis of alternating direction method of multipliers
  for a family of nonconvex problems.
\newblock {\em SIAM Journal on Optimization}, 26(1):337--364, 2016.

\bibitem{kang2012gigatensor}
U.~Kang, E.~Papalexakis, A.~Harpale, and C.~Faloutsos.
\newblock Gigatensor: scaling tensor analysis up by 100 times-algorithms and
  discoveries.
\newblock In {\em KDD12}, pages 316--324. ACM, 2012.

\bibitem{ykim16trip}
Y.~Kim, J.~Sun, H.~Yu, and X.~Jiang.
\newblock {\em Federated Tensor Factorization for Computational Phenotyping}.
\newblock http://dm.postech.ac.kr/trip, 2017.

\bibitem{kissner2005privacy}
L.~Kissner and D.~Song.
\newblock Privacy-preserving set operations.
\newblock In {\em Advances in Cryptology--CRYPTO 2005}, pages 241--257.
  Springer, 2005.

\bibitem{Kononenko:2001up}
I.~Kononenko.
\newblock {Machine learning for medical diagnosis: history, state of the art
  and perspective.}
\newblock {\em Artificial Intelligence in Medicine}, 23(1):89--109, Aug. 2001.

\bibitem{kuhn1955hungarian}
H.~W. Kuhn.
\newblock The hungarian method for the assignment problem.
\newblock {\em Naval research logistics quarterly}, 2(1-2):83--97, 1955.

\bibitem{Ledley:1991wc}
R.~S. Ledley and L.~B. Lusted.
\newblock {Reasoning foundations of medical diagnosis}.
\newblock {\em M.D. computing : computers in medical practice}, 8(5):300--315,
  Sept. 1991.

\bibitem{li2015vertical}
Y.~Li, X.~Jiang, S.~Wang, H.~Xiong, and L.~Ohno-Machado.
\newblock Vertical grid logistic regression (vertigo).
\newblock {\em Journal of the American Medical Informatics Association}, page
  ocv146, 2015.

\bibitem{lin2011linearized}
Z.~Lin, R.~Liu, and Z.~Su.
\newblock Linearized alternating direction method with adaptive penalty for
  low-rank representation.
\newblock In {\em Advances in neural information processing systems}, pages
  612--620, 2011.

\bibitem{murphy2010serving}
S.~N. Murphy, G.~Weber, M.~Mendis, V.~Gainer, H.~C. Chueh, S.~Churchill, and
  I.~Kohane.
\newblock {Serving the enterprise and beyond with informatics for integrating
  biology and the bedside (i2b2)}.
\newblock {\em Journal of the American Medical Informatics Association},
  17(2):124--130, 2010.

\bibitem{PCORnetDataPrivacyTaskForce2015}
{PCORnet Data Privacy Task Force}.
\newblock {Technical Approaches for Protecting Privacy in the PCORnet
  Distributed Research}, 2015.

\bibitem{MIMICII}
M.~Saeed, M.~Villarroel, A.~T. Reisner, G.~Clifford, L.-W. Lehman, G.~Moody,
  T.~Heldt, T.~H. Kyaw, B.~Moody, and R.~G. Mark.
\newblock Multiparameter intelligent monitoring in intensive care ii
  (mimic-ii): A public-access intensive care unit database.
\newblock {\em Critical Care Medicine}, 39:952--960, May 2011.

\bibitem{vaidya2008privacy}
J.~Vaidya, H.~Yu, and X.~Jiang.
\newblock Privacy-preserving svm classification.
\newblock {\em Knowledge and Information Systems}, 14(2):161--178, 2008.

\bibitem{wang2015rubik}
Y.~Wang, R.~Chen, J.~Ghosh, J.~C. Denny, A.~Kho, Y.~Chen, B.~A. Malin, and
  J.~Sun.
\newblock Rubik: Knowledge guided tensor factorization and completion for
  health data analytics.
\newblock In {\em KDD15}, pages 1265--1274. ACM, 2015.

\end{thebibliography}

\end{document}